\documentclass[runningheads]{llncs}

\usepackage[T1]{fontenc}
\usepackage{graphicx}
\begin{document}
\title{Drone navigation and license place detection for vehicle location in indoor spaces
%\thanks{Supported by organization x.}
}
\author{Moa Arvidsson\inst{1} \and
Sithichot Sawirot\inst{1} \and
Cristofer Englund\inst{1} \and \\
Fernando Alonso-Fernandez
\inst{1} \and
Martin Torstensson\inst{2} \and
Boris Duran\inst{2}}
\authorrunning{M. Arvidsson et al.}
\institute{School of Information Technology, Halmstad University, Sweden \\ \email{cristofer.englund@hh.se, feralo@hh.se} \and RISE Viktoria, Gothenburg \\ \email{martin.torstensson@ri.se, boris.duran@ri.se}}
\maketitle              % typeset the header of the contribution
\begin{abstract}

Millions of vehicles are transported every year, tightly parked in vessels or boats.
To reduce the risks of associated safety issues like fires, knowing the location of vehicles is essential, since different vehicles may need different mitigation measures, e.g. electric cars.
This work is aimed at creating a solution based on a nano-drone that navigates across rows of parked vehicles and detects their license plates.
We do so via a wall-following algorithm, and a CNN trained to detect license plates.
All computations are done in real-time on the drone, which just sends position and detected images that allow the creation of a 2D map with the position of the plates.
Our solution is capable of reading all plates across eight test cases (with several rows of plates, different drone speeds, or low light) by aggregation of measurements across several drone journeys.

\end{abstract}

\section{Introduction}

The business of transporting vehicles is constantly expanding. Millions of cars are transported in different ways, such as by truck, air, rail, or vessel \cite{1}.
The most cost-effective method is by boat \cite{2}.
Today there are ocean vessels built to carry up to 8000 vehicles.
There are currently about 1400 vessels globally \cite{3}, and an estimated 7 million cars carried on vessels around the world every year.

Due to the high density of packed vehicles on decks, finding and identifying specific ones can be challenging.
The mixed storage of combustion engines and electric vehicles or vehicles of different sizes further complicates the situation.
Accurate knowledge of vehicle locations is crucial for safety reasons, such as in the event of a fire, as different measures are needed with electric vehicle batteries.

%Vehicles are packed very closely at each deck to fit as many as possible. Therefore, it is challenging to traverse through them to identify or find a car.
%
%In some cases, vehicles of different sizes are parked together, creating an uneven path between rows.
%
%Also, both combustion engine and electric cars are shipped together.
%
%One of the greatest risks is fire.
%
%In such contexts, knowing exactly where vehicles are parked can reduce the risk e.g. in case of fire, and help with the mitigation measures.
%
%For example, fires in batteries of electric vehicles should not be fought with water.

A simple way to identify them is to detect the license plate or identification number.
This is possible via CCTV cameras, but plates are usually small and maybe obstructed due to tightly parked vehicles.
Therefore, a solution based on a nano drone is investigated, since
it can fit in narrow spaces. An onboard camera can carry out plate detection simultaneously.
The proposed solution uses a wall-following strategy for navigation, treating rows of packed vehicles as walls.
%
%Simultaneously, the drone detects plates by object classification.
The images and drone position are sent to a remote client, which builds a 2D map that depicts the drone's path and detection results.
This solution offers a promising method for efficiently identifying vehicles in crowded storage areas.

\begin{figure}
\centering
\includegraphics[width=0.65\textwidth]{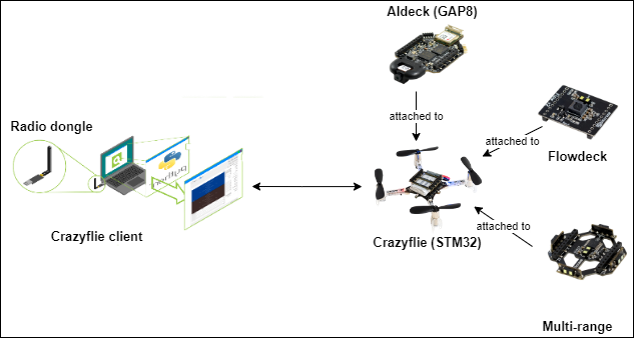}
\caption{System overview.} \label{fig:system_overview}
\end{figure}

\section{Related Works}

We describe existing methods for navigation and object detection with drones.
One gap is the limited size, weight, and computation onboard, making the use of predominant deep learning solutions a challenge \cite{8}.

\subsection{Navigation}

Unmanned Aerial Vehicles (UAVs) or drones require a navigation system to determine their position and trajectory. %, and navigate their surroundings.
%
%Range sensors can also aid by detecting obstacles.
%
GPS navigation is feasible outdoors, but not indoors.
%
%Outdoors, navigation can be achieved via GPS, but it is unfeasible indoors.
%
An indoor positioning system like Bitcraze's Loco Positioning system can be employed \cite{17}.
It includes anchors, similar to GPS satellites, and a tag that acts as a receiver.
It provides absolute positioning in 3D with a range of 10 meters.
However, in large areas like ship decks, with dozens/hundreds of meters, equipping the entire space with anchors becomes costly.

When GPS or tags are not available, cameras can be used to navigate unknown spaces.
Simultaneous Localization and Mapping (SLAM) is a well-established technique, researched for many years \cite{7747236}.
%
%SLAM (Simultaneous Localization and Mapping) is the foundation of image-based self-navigation. %, allowing to localize (L) and map (M) the environment.
%
%The field accumulates decades of research \cite{7747236}.
%
The first SLAM on a small drone was achieved with a residual Convolutional Neural Network (CNN) called DroNet \cite{8264734}, followed by PULP-DroNet \cite{9606685}, which enabled onboard computation on a nano drone like ours.
%
%The first SLAM implementation on a small drone was a residual Convolutional Neural Network (CNN) DroNet \cite{8264734} and its sequel PULP-DroNet \cite{9606685}, the latter being capable of using onboard computation of a nano drone like ours.
%the CrazyFlie without the aid of base stations. To achieve that, intensive CNN optimization and pruning was done, including fixed-point representation, quantization of layer weights, and manual tuning of the architecture.
%
However, they only provide collision probability and recommended steering angle to avoid collisions. Additionally, they are trained with outdoor data from car driving, as they are designed for autonomous navigation on streets.
%
%Major limitations of DroNet and PULP-Dronet are that they just provide the collision probability and recommended steering angle to avoid it, and that they are trained with outdoor data from car driving, since they are designed to navigate autonomously through streets.
%
%
For indoor environments, the swarm gradient bug algorithm (SGBA) \cite{24} was proposed.
It is a minimal solution to explore an unknown environment autonomously using a 'wall-following' behavior. %, which was demonstrated to be able to explore unfamiliar environments.
Unlike SLAM, SGBA requires less processing power, making it more suitable for our requirements.
A row of vehicles can be considered walls, with the drone following along with the camera facing them while scanning for plates. This simplifies the navigation while still achieving the goal of identifying vehicles in a crowded storage area.

\subsection{Object Detection}

For object detection, the state-of-the-art is given by region proposal networks (RPN), such as region-based Convolutional Neural Networks (R-CNN) \cite{10}, or Single-Shot Detection networks (SSD), such as YOLO \cite{7780460}. RPNs require two stages, one for generating region proposals, and another for object detection. SSDs predict position and object type in a single stage, making them faster and more efficient, at the cost of less accuracy. However, the size of the networks behind any of these models (e.g. Darknet or EfficientNet) is too large for a nano drone.
%
%
%
%Object detection can be achieved by region proposal networks (RPN), such as region-based Convolutional Neural Networks (R-CNN) \cite{10}.
%
%%, Region-based Fully Convolu- tional networks (R-FCN), and Feature Pyramid Networks (FPN).
%
%
%However, R-CNNs have the drawback of requiring two stages: one for generating region proposals and another for object detection.
%
%%One disadvantage of R-CNNs is that they need two stages, one for generating region proposals, and another for detecting the object of each proposal.
%
%%It has three stages (region proposal, feature extraction, and classification) and employs a sliding-window approach, which has a high computational cost.
%
%In contrast, Single-Shot Detection networks (SSD) predict position and object type in a single stage, making them faster and more efficient.
%
%The state-of-the-art in SSD is given by YOLO (You Only Look Once) networks \cite{7780460}, with the latest version being YOLO v7.
%
%The underlying CNN backbone varies across versions, such as Darknet or EfficientNet.
%
%%Darknet-19 (v2), Darknet-53 (v3), CSPNet (v4), EfficientNet (v5),
%
%However, the size of these models is too large for a nano drone.
%
To address this, Greenwaves Technologies, the manufacturer of the GAP8 processor used by our drone, offers several SSD classification CNNs based on different architectures of the much lighter MobileNet \cite{14}.

Detection of objects such as vehicles, people, fruits, pests, etc., from UAVs, is gaining attention \cite{RAMACHANDRAN2021215}.
However, it mostly involves drones of bigger size than ours.
Another difference is that studies mostly use aerial images taken from a certain height and with the objects appearing small compared to the background.
Here, the problem is reversed. The UAV will fly relatively close to the target object, making that, for example, vehicles do not fit entirely into the image.

\begin{figure}[t]
\centering
\includegraphics[width=0.55\textwidth]{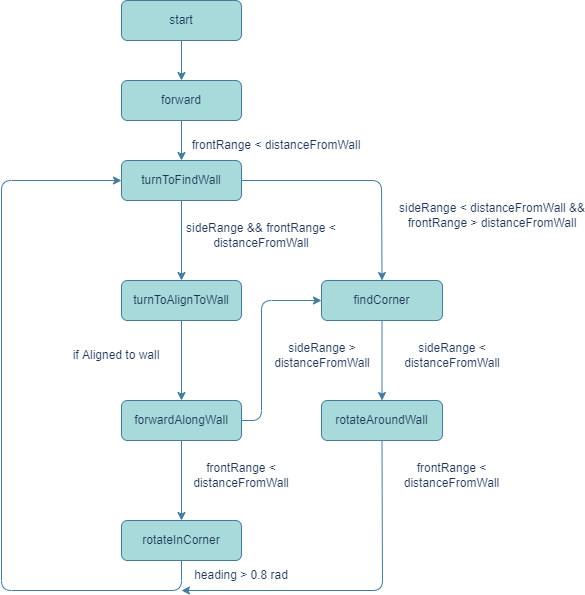}
\caption{State diagram of the wall-following algorithm.} \label{fig:fig6_wall_following}
\end{figure}

\section{Methodology}

%\subsection{System Overview}
%\label{sect:system}

%The system has several components, shown in Figure~\ref{fig:system_overview}, for autonomous flight, detection and mapping.

\subsection{Hardware}
\label{sect:HW}

The system has several components (Figure~\ref{fig:system_overview}).
At the core is the \textbf{CrazyFlie drone}, which utilizes an MCU (STM32) for autonomous flight control, position estimation, sensor data collection, and communication with other components.
The CrazyFlie, manufactured by Bitcraze, is a nano quadrotor with a small 10 cm wingspan, classified as nano due to its small size and low power.
Weighing 27 grams, it features low latency and long-range radio capabilities, Bluetooth LE, onboard charging via USB, and expansion decks for additional features.
%
%The drone is equipped with an STM32F405 main application MCU and an IMU.
%
The flight time is 7 min, and the max recommended payload is 15g.
%
%According to the developers during Bitcraze's yearly workshop, a 5g increase in weight may limit the flight time to 4 minutes instead of 7 minutes.
%
Bitcraze offers a range of expansions, with the relevant ones for this research described next.
%
%The CrazyFlie is connected with two sensor expansion decks, the MultiRanger deck and the Flow deck. The MultiRanger deck detects any object around the CrazyFlie, while the Flow deck keeps track of the drone’s movements.

An \textbf{AI deck} (of weight 4.4g) allows for AI computations onboard to manage, for example, autonomous navigation.
It is equipped with a GreenWaves Technologies GAP8 system-on-chip processor \cite{13}, featuring a CNN accelerator optimized for image and audio algorithms. The AI deck also includes I/O peripherals to support devices such as cameras and microphones.
Additionally, an integrated ESP32 chip provides WiFi connectivity to stream images from a grayscale ultra-low-power Himax camera.
%
%The main advantage of using the Greenwave chip is that it reduces deployment and operating costs.
%
%An AI deck is also connected to the CrazyFlie with its own MCU (GAP8), where the classification is run.
%
The AI deck sends the computation result to the CrazyFlie, which relays the information along with the drone’s estimated position to the CrazyFlie client’s console via radio.

A \textbf{Flow deck} keeps track of the drone’s movements. A VL53L1x ToF sensor measures the distance to the ground and a PMW3901 optical flow sensor measures ground movement. These sensors allow the CrazyFlie to be programmed to fly distances in any direction or hover at a certain altitude. The flow deck can measure up to 4 meters and weights 1.6g. %The optical flow and height measurements are fused in the Extended Kalman Filter of the CrazyFlie firmware.

A \textbf{MultiRanger deck}, of weight 2.3g, detects any object around the CrazyFlie. It measures the distance to objects in 5 directions: front, left, right, back, and up. The maximum distance is 4 meters with millimeter precision.

\begin{figure}[t]
\centering
\includegraphics[width=0.75\textwidth]{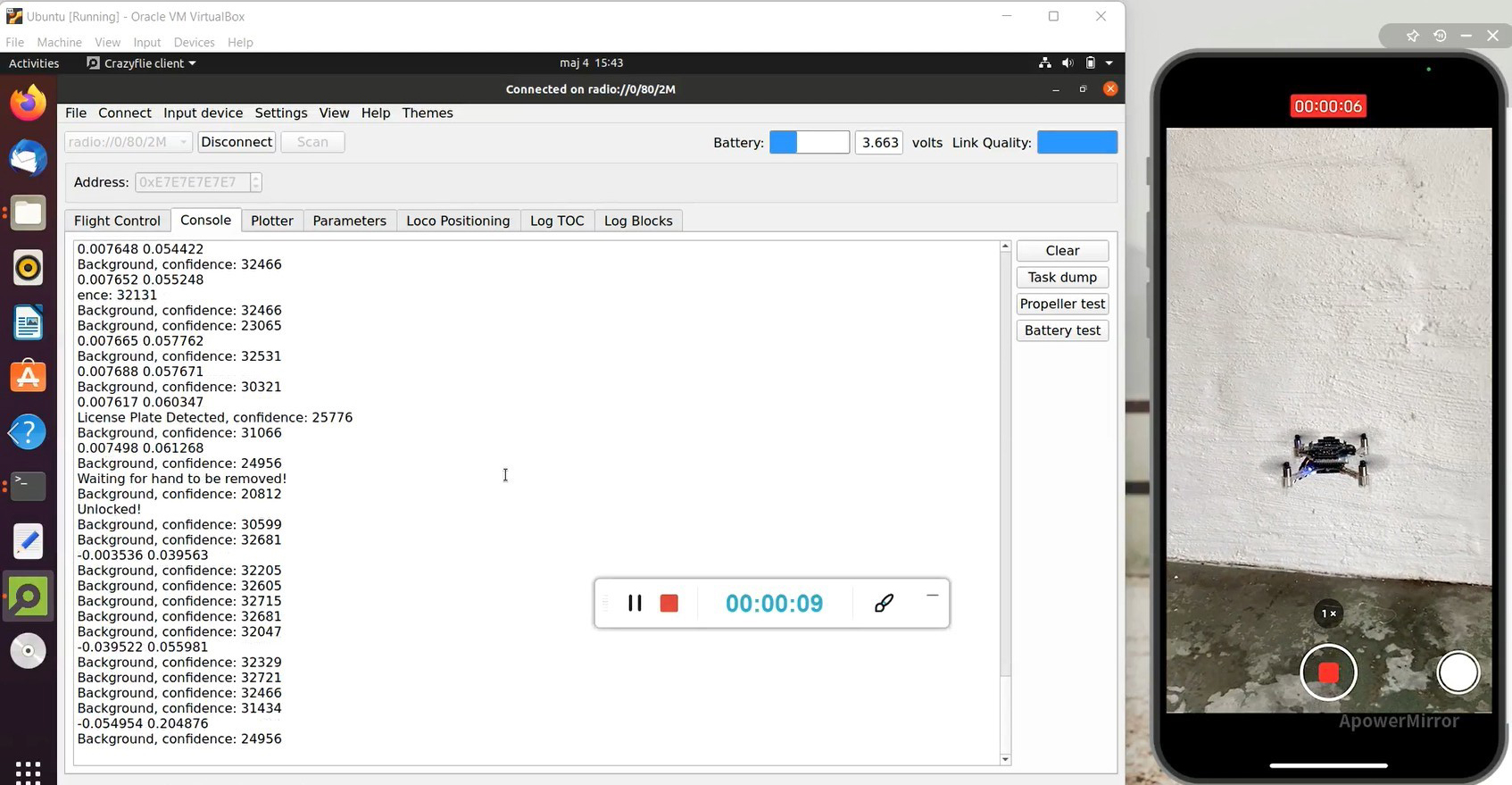}
\caption{Screenshot of system recording.} \label{fig:fig23_system_screenshot}
\end{figure}

\begin{figure}[t]
\centering
\includegraphics[width=0.75\textwidth]{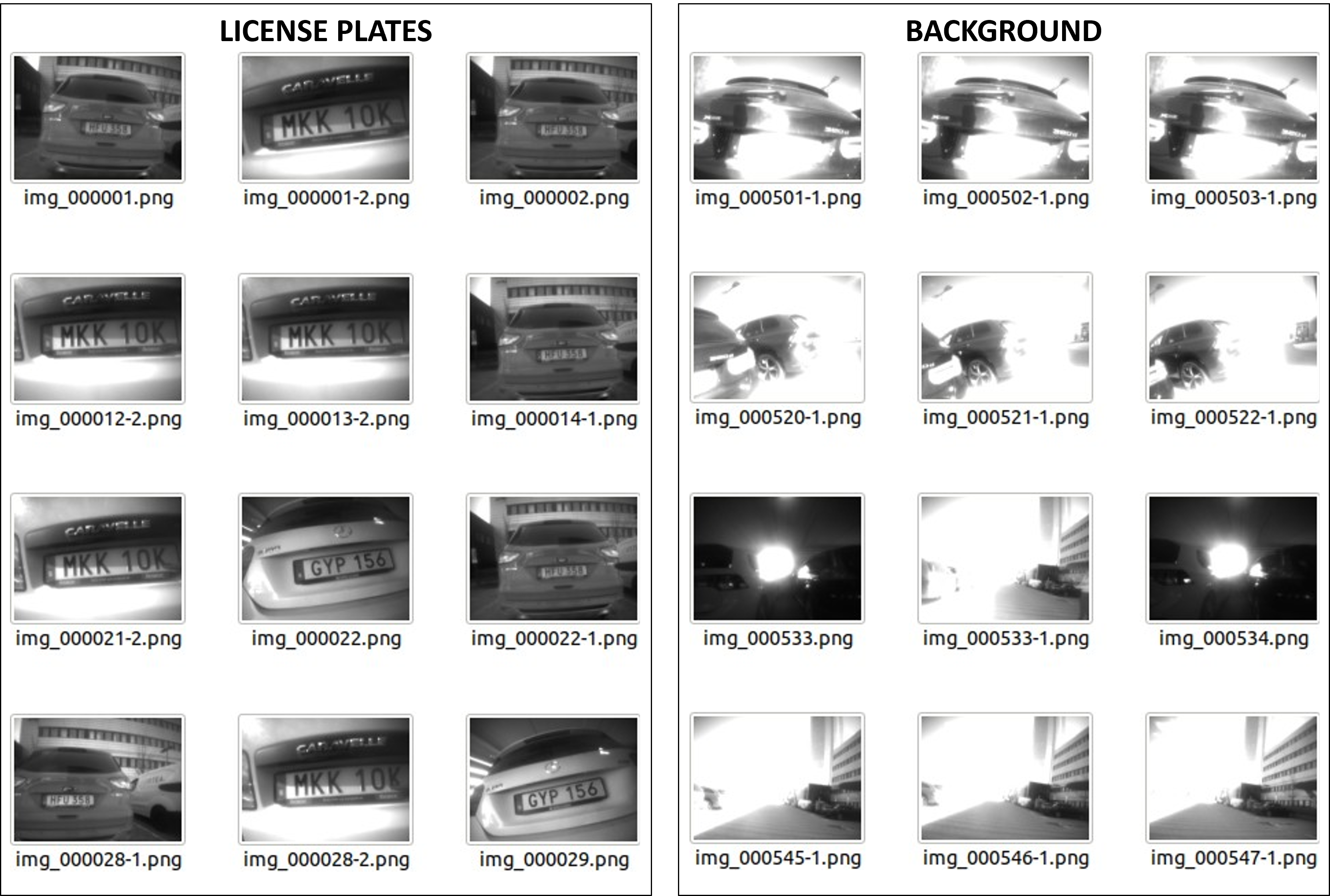}
\caption{Collected images of license plates and background with HIMAX camera.} \label{fig:fig15_16_database}
\end{figure}

\subsection{Software and Data}
\label{sect:SW}

When creating a system that navigates and detects objects onboard a nano-drone, the computing capabilities can be a challenge. The software components of our system are divided into three tasks: navigation, detection, and mapping, which are explained next.

Considering a row of parked cars as a wall due to their tight packing, we use the SGBA's wall-following algorithm \cite{24} as a lighter and more power-efficient \textbf{navigation} solution compared to SLAM.
The original algorithm was modified to have the drone face the walls while flying sideways, allowing the camera to scan for license plates.
The MultiRanger deck is used to detect obstacles, while the Flow deck helps maintain the drone's stability at a specific height and measures the distance and direction of movement.
The implemented wall-following algorithm, shown in Figure~\ref{fig:fig6_wall_following}, involves the drone moving forward until a wall is detected in front.
It then aligns itself with the wall and continues moving forward along it. If the side range sensor loses sight of the wall, the drone seeks a corner and rotates around it.
The drone then aligns itself to the new wall and moves forward along it.
If it finds a wall from the front range sensor instead, it will rotate in the corner, align with the new wall, and move along it.

For \textbf{detection}, a binary approach is followed, where the system outputs either 'license plate' or 'background' as a result.
Real-time computation is crucial since the drone needs to simultaneously fly and perform classification.
To meet the computational constraints, MobileNet v2 is chosen as the classification backbone, as it reduces complexity costs and network size compared to other models, making it suitable for mobile devices like drones \cite{28}.
Bitcraze provides a demo for classifying Christmas packets \cite{29} using the AI deck of the CrazyFlie, which serves as the basis for this research, although it needs adaptation to our particular scenario and data.
%
%The size of MobileNet V2 is about 14MB, which is about what the AI deck can handle.
%
The provided model is trained for a different task (Christmas packets) and their data is collected from just one position.
%
%The detection model provided by Bitcraze \cite{29} is trained on a different scenario than ours (Christmas packets), and their data is collected from just one position.
%
To use it for our intended purposes, we captured our own customized \textbf{database} of 'license plates' and 'background' (Figure~\ref{fig:fig15_16_database}) with the HIMAX camera attached to the AI deck.
The dataset consists of training/validation images with 747/180 license plate samples and 743/183 backgrounds, all grayscale and at a resolution of 320$\times$320.
The detection model is trained using the captured samples for 100 epochs.
To deploy the deep learning model on the GAP8 processor, we use the GAP flow tool provided by GreenWaves Technologies inside the GAP8 SDK \cite{13}.

\textbf{Mapping} plays a crucial role in monitoring the target objects and keeping track of their positions.
This is achieved by sending the drone position and the classification result to a remote client console.
When navigating, the CrazyFlie captures images with the HIMAX camera at 2 Hz, which are classified in real-time, generating a confidence score for each. The results, along with the drone's current position, are transmitted to a remote client (Figure~\ref{fig:fig23_system_screenshot}).
The client connects to the CrazyFlie using a USB radio dongle. This allows the client to display the classification results, drone position, and path on a scaled figure. The classification results can be color-coded for easier interpretation (e.g. Figure~\ref{fig:figures_results_T1_T4}).

%\subsection{Data}
%\label{sect:data}

\section{Experiments and Results}

To evaluate the system, precision (P) and recall (R) are used to measure the classification performance.
Precision measures the proportion of positive classifications (said to be a license plate), which are actually an image with a license plate.
Recall, on the other hand, measures the proportion of actual positive cases (images with a license plate) which are correctly classified as positive (said to have a license plate).
A summarizing metric is the F1-score, the harmonic mean of P and R, computed as F1=2*(P*R)/(P+R).

\begin{figure}[t]
\centering
\includegraphics[width=0.85\textwidth]{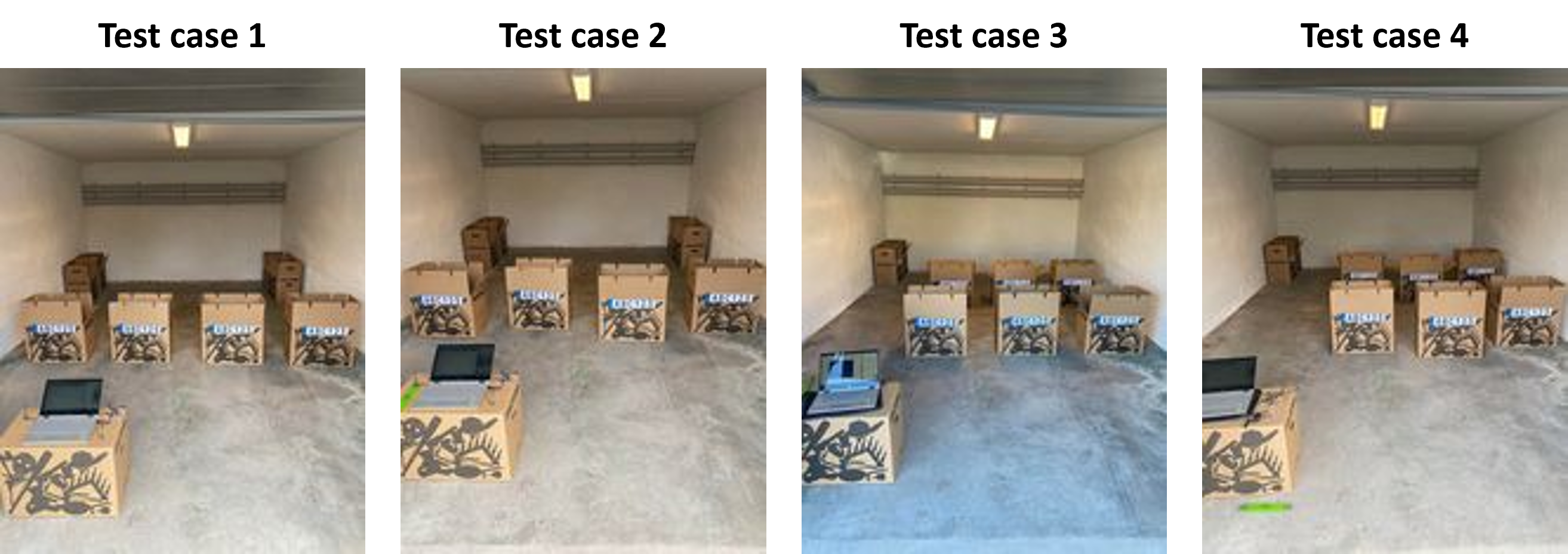}
\caption{Setup of different flight test cases (Section~\ref{sect:test_cases}).} \label{fig:fig18_21_test_cases}
\end{figure}

\subsection{Test Cases}
\label{sect:test_cases}

To simulate vehicles, we employ moving boxes with printed-out standard-size license plates (Figure~\ref{fig:fig18_21_test_cases}) on a garage of size 6 $\times$ 3.6 meters.
A fixed start point will be used so that the drone starts in the same position every time.
We define four different setups for testing with different paths and placements of the objects. Each case will be evaluated three times (i.e. the drone will be deployed on three different occasions).
The test cases are set up to evaluate the performance of the wall following algorithm and license plate detection, including when the drone has to proceed across several walls.
We also aim at evaluating situations where the objects with license plates are not in a straight line.
The following test cases are thus considered, shown in Figure~\ref{fig:fig18_21_test_cases}:

\begin{enumerate}

    \item One row with 4 moving boxes in a straight line, with a gap of 25 cm between each. The row is 2 m away from the starting navigation point.

    \item The same previous setup, but the boxes were not placed in a straight line.

    \item Two rows with 3 moving boxes each in a straight line and separated 25 cm, leaving space for the drone to turn around. The first row is 2 m away from the starting navigation point, and the second is 1.37 m behind the first one.

    \item The same previous setup, but the boxes were not placed in a straight line.

\end{enumerate}

\begin{figure}[t]
\centering
\includegraphics[width=0.97\textwidth]{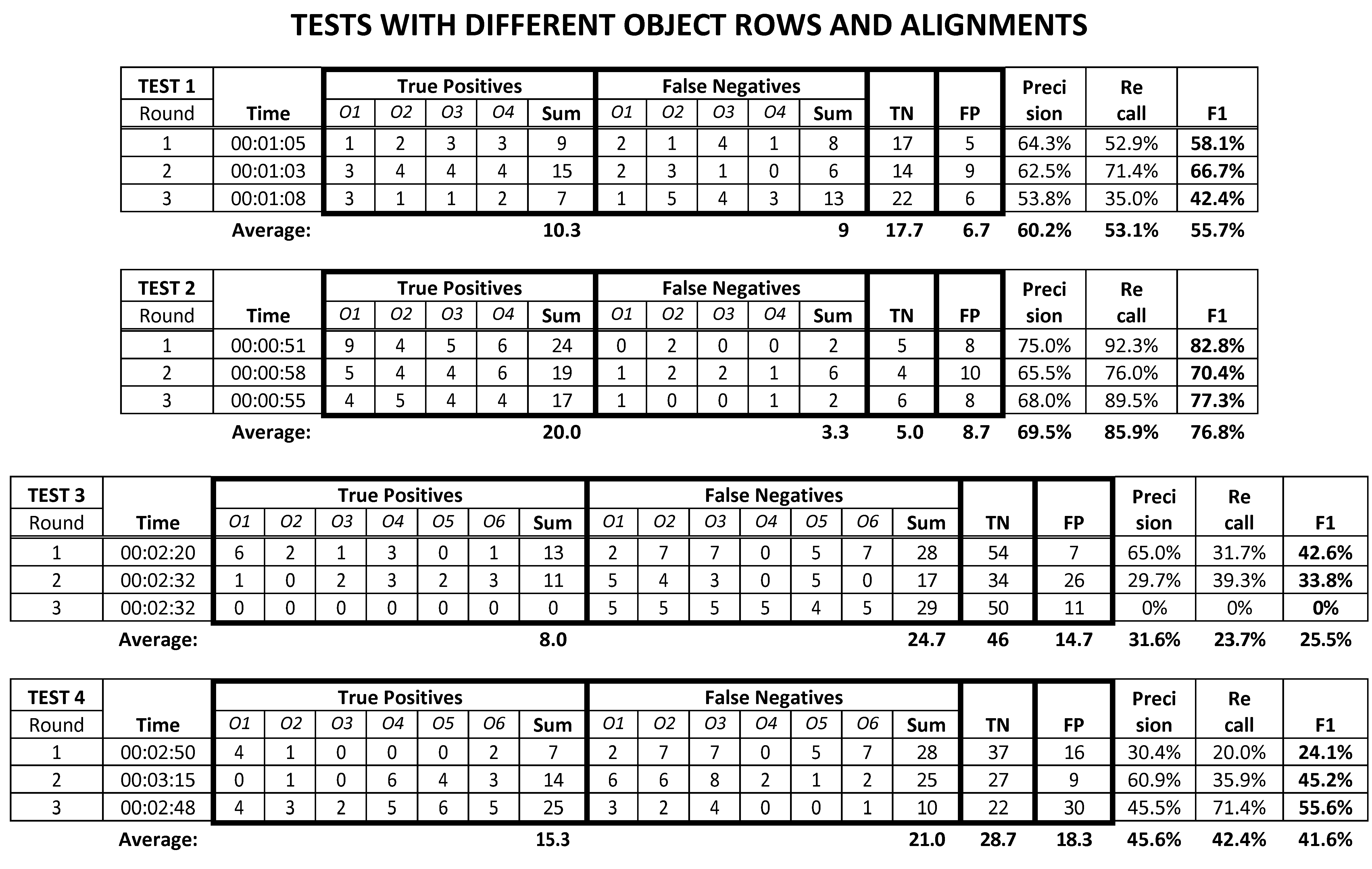}
\caption{Classification results and flying time of different test cases (Section~\ref{sect:test_cases}).} \label{fig:tables_results_T1_T4}
\end{figure}

\begin{figure}[t]
\centering
\includegraphics[width=0.95\textwidth]{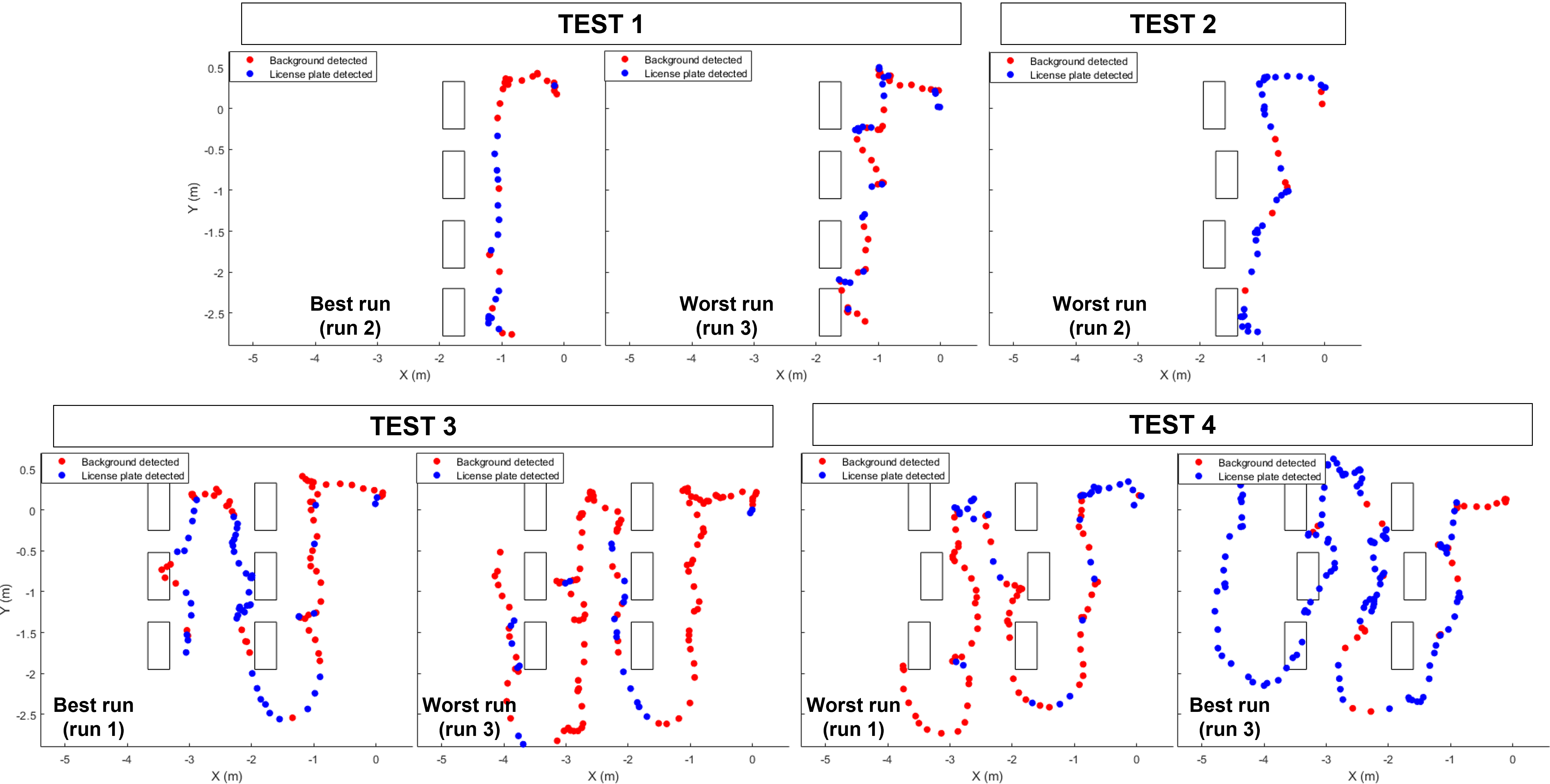}
\caption{Drone paths on different flight test cases (Section~\ref{sect:test_cases}).} \label{fig:figures_results_T1_T4}
\end{figure}

Figure~\ref{fig:tables_results_T1_T4} gives the classification results and the flying time of each round across the different tests cases, whereas Figure~\ref{fig:figures_results_T1_T4} shows the 2D grid with the drone path and classification output of selected rounds during the journey (the worst and the best round, based on the F1-score).

Since the drone classifies continuously, there are more true positives (TP) than objects, because the same object is captured in different frames.
TP indicates how many true 'license plates detected' the drone prints out when flying past the boxes.
For example, in test case 1, with four objects tested, there are between 7 and 15 true positives (depending on the round).
It can also be seen that in all tests, all objects are detected at least once in some of the rounds.
This means that all license plates are captured if the drone is allowed to do two or three rounds of navigation. At the remote client, the actual plate number could be extracted (not implemented in this work), making possible to consolidate the different true positives of the same number into one single instance.
These results indicate that the implemented system is able to work across the different layouts and object alignment tested.

The system also shows false negatives (FN), meaning that the drone classifies as 'background' an image containing a license plate.
On average, the false negatives are less than the true positives in the single-row experiments (tests 1 and 2), but in the two-rows experiments (tests 3 and 4), it is the opposite.
In tests 3 and 4, there are more objects to classify (six vs. four) and the drone is navigating more time, which obviously results in more available images with positives.
But having to navigate and deal with wall/row corners (Figure~\ref{fig:figures_results_T1_T4}) may produce many of those images showing a license plate from a very difficult perspective, impacting the capability to detect them.
However, based on the previous considerations about the true positives, this should not be an issue because the drone is able to capture all license plates in several images across different journeys.
The system is not free of false positives (FP) either (i.e. background frames said to have a license plate), an issue that could also be resolved at the remote client with a more powerful classifier that concentrates only on the selected frames sent by the drone and discards erroneous cases.
The number of true negatives (TN) is also usually higher than the false positives, meaning that a high proportion of background images are labeled correctly.

When analyzing if boxes are in a straight line or not (case 1 vs. 2, and 3 vs. 4), it is interesting to observe that the performance is better when they are not forming a straight line (observe P, R, and F1-score).
The drone does not seem to have difficulties in following the 'wall' of boxes even if they are not completely aligned, as seen in the paths of Figure~\ref{fig:figures_results_T1_T4}, and this indeed produces better detection results overall.
On the other hand, the two-row tests (cases 3 and 4) show worse performance overall than the single-row tests (cases 1 and 2), an issue that could be attributed to the mentioned imaging perspective of the two-rows navigation that causes a greater amount of false negatives and false positives.
Also, the flight time obviously differs. Two rows demand extra time for the drone to turn around, find the way and navigate across a bigger amount of objects.
%
%The flight time of cases 1-2 (one row of objects) is obviously less than cases 3-4, which have two rows, thus demanding extra time for the drone to turn around and find itself.
%
As seen in Figure~\ref{fig:figures_results_T1_T4}, the drone navigates each row of boxes on the two sides, as expected from the wall following algorithm, which increases flying time to well beyond double.
%
%In our case, only one side of the boxes contain license plates. ??

When analyzing the drone paths in Figure~\ref{fig:figures_results_T1_T4}, it can be seen that in several cases, the drone has difficulties flying following a straight line.
This is, very likely, a limitation of employing a wall-following algorithm, since we use objects that have some gap among them (25 cm). Sometimes, the drone has a tendency to move toward the gap between the boxes, although it is not always the case.
In the two-row setup, it is capable of moving along wall corners and row ends without issues.
Together with the fact that the system does not miss any license plate if several rounds are allowed, these results validate our overall approach.
In a few runs, the paths show that the drone flies over the boxes, but it did not happen in any run. The drone was flying correctly, close to the boxes, but it seems that it was estimating its position incorrectly.

%The latter it is not expected to happen with real vehicles due to a higher height that would be detected by the MultiRanger deck.
%

\begin{figure}[t]
\centering
\includegraphics[width=0.97\textwidth]{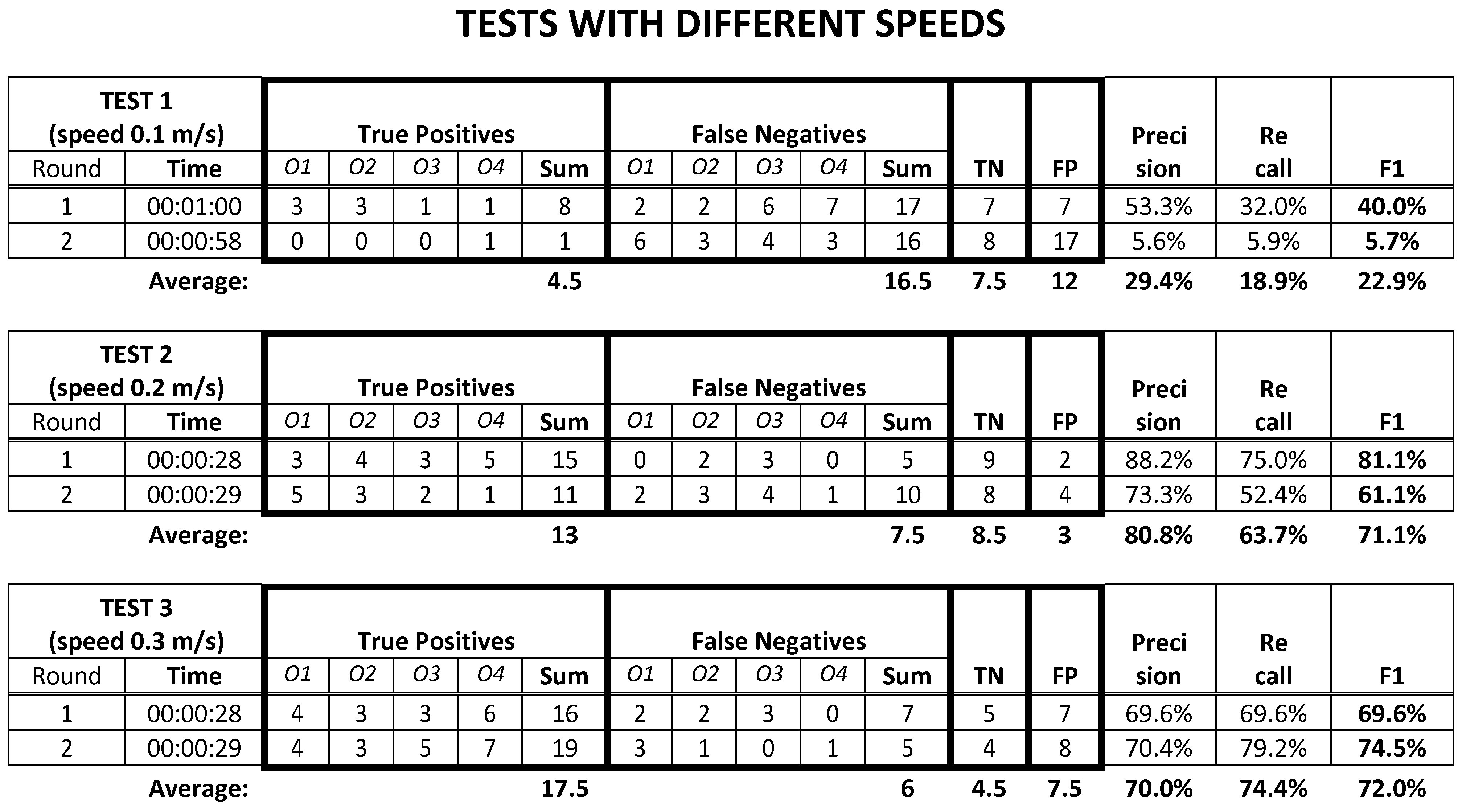}
\caption{Classification results and flying time of speed test cases (Section~\ref{sect:speed_factor}).} \label{fig:speed_factor}
\end{figure}

\subsection{Speed Factor}
\label{sect:speed_factor}

Three different speeds of the drone (0.1, 0.2, and 0.3 m/s) have also been tested across two rounds per speed to see how it impacts performance.
This experiment is carried out on scenario 1 of the previous sub-section (a single row of 4 boxes in a straight line).
Figure~\ref{fig:speed_factor} gives the classification results and the flying time of each round.

Also, here, all license plates are captured across the two rounds, regardless of the speed.
An interesting result is that the worst results are given at the slowest speed, with the system missing many plates as false negatives, and producing many false positives as well.
At 0.2 or 0.3 m/s, the amount of false negatives and false positives is significantly less, with a conversely higher amount of true positives.
Comparatively, a higher speed does not imply worse results in general. This could be exploited to complete the expedition faster, counteracting the battery issue mentioned in the previous sub-section.

%Also, at 0.2 or 0.3 m/s, the average F1-score is approximately equal, but the two cases differ in precision and recall.
%
%At 0.2 m/s, a bigger precision is generally obtained (the majority of positives contain a license plate),
%
%whereas at 0.3 m/s, a bigger recall is obtained ().

\begin{figure}[t]
\centering
\includegraphics[width=0.4\textwidth]{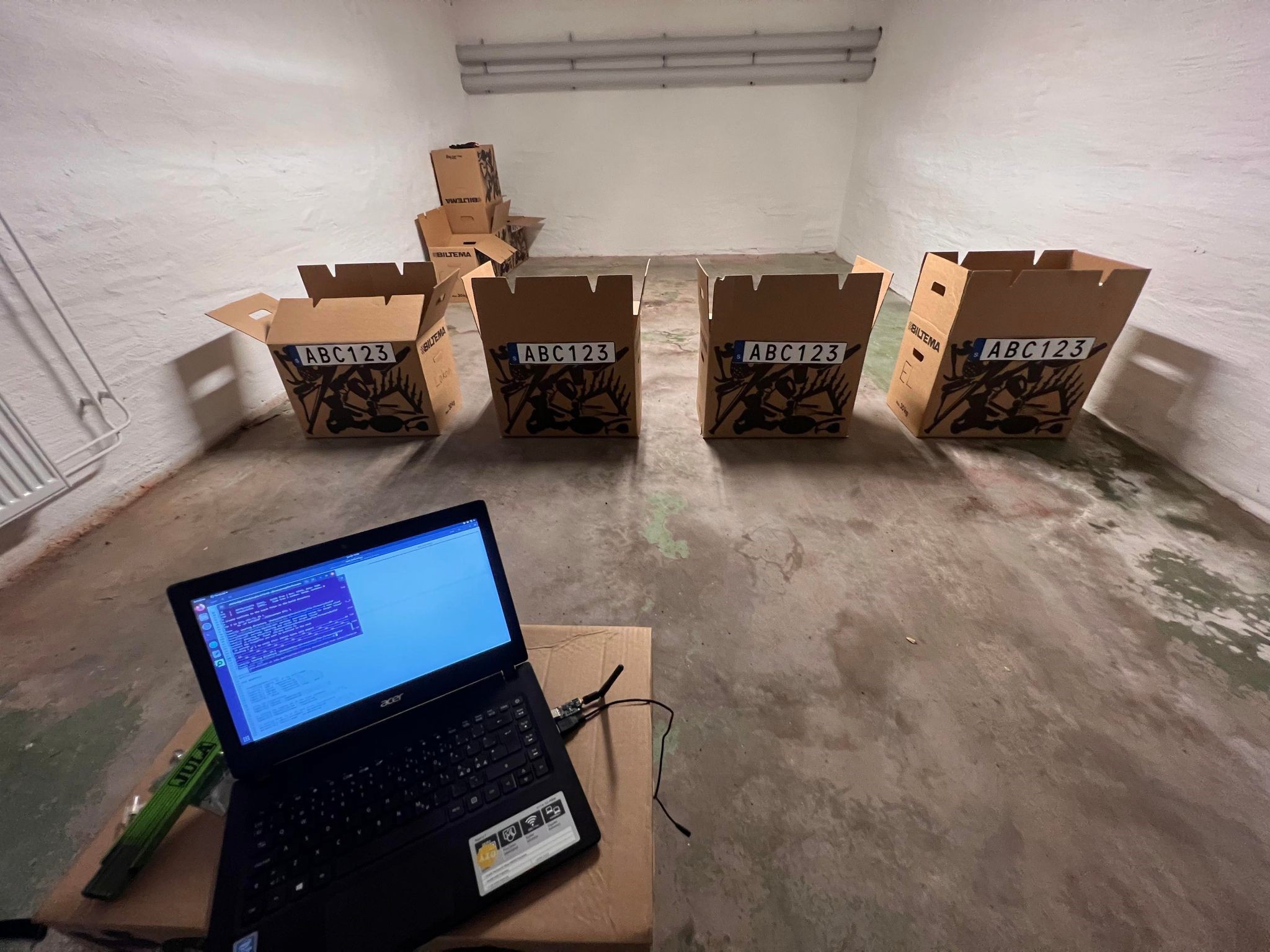}
\caption{Setup of the low light test case (garage door closed) (Section~\ref{sect:light_factor}).} \label{fig:fig22_low_light_case}
\end{figure}

\begin{figure}[t]
\centering
\includegraphics[width=0.97\textwidth]{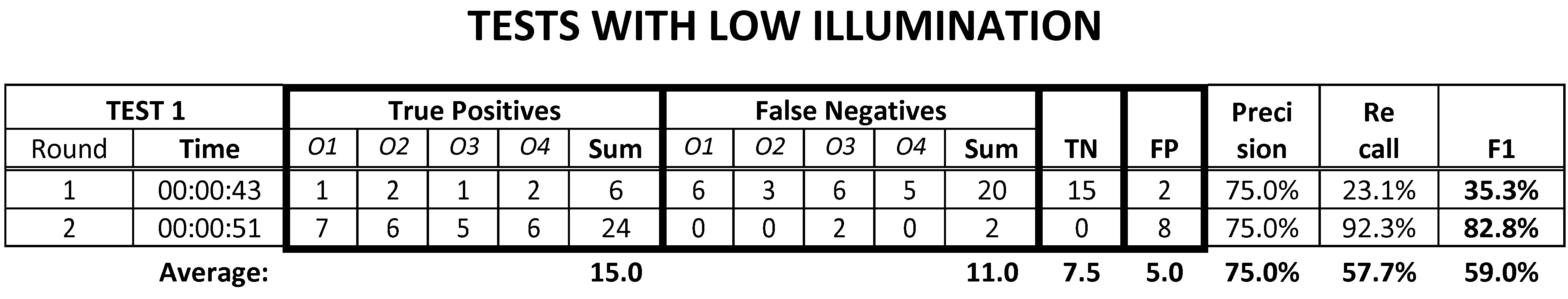}
\caption{Classification results and flying time of low light test cases (Section~\ref{sect:light_factor}).} \label{fig:light_factor}
\end{figure}

\subsection{Light Factor}
\label{sect:light_factor}

This test was conducted with the garage door closed, so the environment is darker and only illuminated by some ceiling lamps (Figure~\ref{fig:fig22_low_light_case}).
The test is done over two rounds with a single row of 4 boxes in a straight line.
Figure~\ref{fig:light_factor} gives the results and the flying time of each round.
As in the previous cases, all license plates are captured across the two rounds, so the evaluated light conditions do not have an impact on the detection either.
One possible effect of the lower light is the dispar results between the two rounds in detecting plates. The first round has many false negatives, whereas the second round has many true positives. Also, in round two, no background is detected correctly (true negatives=0).
%
%The F1-score of Figure~\ref{fig:light_factor} is on par with the same test of Figure~\ref{fig:tables_results_T1_T4} (59\%).
%
It must be stated as well that the navigation is not be affected by darkness, since the sensors of the Flow and MultiRanger decks are not based on visible illumination.
Only the classification would need cameras and software adaptation capable of working in very low light conditions, such as infrared cameras.

\section{Conclusions}

This work has presented a system that makes use of a camera on-board a nano drone to navigate across rows of vehicles tightly parked and find their license plate.
We apply a navigation solution based on wall-following, and a MobileNet-based CNN trained to detect license plates. The solution is fully executed onboard a CrazyFlie drone, with just 10 cm wingspan.
The drone position and images are used on a remote client to build a 2D map of the path and license plates detected without relying on positioning systems (e.g. GPS) or tagging.
Our application scenario is transportation, where vehicles are packed closely together, and knowing the exact position of each one and its features (e.g. electric or combustion) can help to mitigate security issues, such as fires.

We have carried out several tests simulating objects with license plates.
Different scenarios are considered, such as several rows (demanding the drone to turn around and find the next row), objects not stacked across a straight line, different drone speeds, or lightning.
In any of them, even if the plates are not detected in every frame, all are captured by aggregation after the drone carries out 2-3 rounds of navigation. This is feasible e.g. on-board vessels after all vehicles have been parked.
The wall-following algorithm, which is less computationally demanding than SLAM \cite{7747236}, correctly navigates across all objects despite a small gap between them. It also works well if the objects are not perfectly aligned.

Our solution assumes that the rows of objects are connected to a wall. It would need extra tweaking if, for example, they do not have a wall on any of their ends.
Also, we only send the drone position and image with a plate detected, but the actual number could be read,
%
%The obtained data would allow to read the exact license plate.
%
either at the drone with additional software or at the remote console.
Processing or sending only images with high detection confidence would allow to save resources at the drone while completing the navigation mission.
The drone path with color codes (Figure~\ref{fig:figures_results_T1_T4}) would also allow obtaining a map with the exact position of each object, as long as at least one image per object is sent eventually.
When several true positives of a license plate are captured, the actual number could be used to group them and filter multiple detections.
In the same manner, checking the extracted number against a list of expected vehicles (a manifesto) would also allow to filter out errors.

Our garage is of size 6 $\times$ 3.6 m and the drone covers it in 1-3 min, depending on the number of rows (Figure~\ref{fig:tables_results_T1_T4}).
The maximum flying time declared by the CrazyFlie is 7 min, so one single charge is able to cover the three rounds of tests 1 and 2, but not of tests 3 and 4. This must be considered when deploying a system like this to a larger space like garages or ships, either by stocking several batteries or more than one drone.
Adding more tasks to the drone itself (e.g. reading the license plate number) would also have an impact on the battery time.
However, if the drone is capable of filtering out several true positives or other errors (at the cost of extra processing), it would transmit fewer images to the remote client, which would reduce battery consumption as a contraposition.

Another possibility is to relieve the drone of detecting plates, and just send the camera stream and position. Detection and number reading would then be done in a more powerful remote client, which could include a larger CNN detector \cite{7780460} for a more precise result.

%Differences of positives/negatives across rounds

%\begin{table}
%\caption{Table captions should be placed above the tables.}\label{tab1}
%\begin{tabular}{|l|l|l|}
%\hline
%Heading level &  Example & Font size and style\\
%\hline
%Title (centered) &  {\Large\bfseries Lecture Notes} & 14 point, bold\\
%1st-level heading &  {\large\bfseries 1 Introduction} & 12 point, bold\\
%2nd-level heading & {\bfseries 2.1 Printing Area} & 10 point, bold\\
%3rd-level heading & {\bfseries Run-in Heading in Bold.} Text follows & 10 point, bold\\
%4th-level heading & {\itshape Lowest Level Heading.} Text follows & 10 point, italic\\
%\hline
%\end{tabular}
%\end{table}

\subsubsection{Acknowledgements.}
This work has been carried out by M. Arvidsson and S. Sawirot in the context of their Master Thesis
at Halmstad University (Computer Science and Engineering).
The authors acknowledge the Swedish Innovation Agency (VINNOVA) for funding their research.
Author F. A.-F. also thanks the Swedish Research Council (VR).

\bibliographystyle{splncs04}
%\bibliography{example}
%

\end{document}